\documentclass[letterpaper, 10 pt, conference]{ieeeconf}

\IEEEoverridecommandlockouts

\usepackage{graphics} 
\usepackage{epsfig} 
\usepackage{mathptmx} 
\usepackage{times} 
\usepackage{amsmath} 
\usepackage{amssymb}  
\usepackage{cite}

\usepackage{amsfonts}
\usepackage{algorithmic}
\usepackage{graphicx}
\usepackage{subcaption}
\usepackage{textcomp}
\usepackage{array}
\usepackage{xcolor}
\usepackage{layouts}
\usepackage{tabularx}
\usepackage{multirow}
\usepackage{gensymb}
\usepackage{booktabs}
\def\BibTeX{{\rm B\kern-.05em{\sc i\kern-.025em b}\kern-.08em
    T\kern-.1667em\lower.7ex\hbox{E}\kern-.125emX}}
    
\title{\LARGE \bf
A Semi-Automated Corner Case Detection and Evaluation Pipeline
}

\author{Isabelle Tülleners$^{1}$, Tobias Moers$^{1}$, Thomas Schulik$^{2}$, Martin Sedlacek$^{2}$  
\thanks{$^{1}$The authors are with the fka GmbH, 52074 Aachen, Germany 
{\tt\small  \{isabelle.tuelleners, tobias.moers\}@fka.de}}
\thanks{$^{2}$The authors are with the ZF Friedrichshafen AG, 88046 Friedrichshafen, Germany
{\tt\small  \{thomas.schulik, martin.sedlacek\}@zf.com}}
}

\begin{document}

\let\oldtwocolumn\twocolumn
\renewcommand\twocolumn[1][]{%
    \oldtwocolumn[{#1}{
    \begin{center}
           \includegraphics[width=\textwidth]{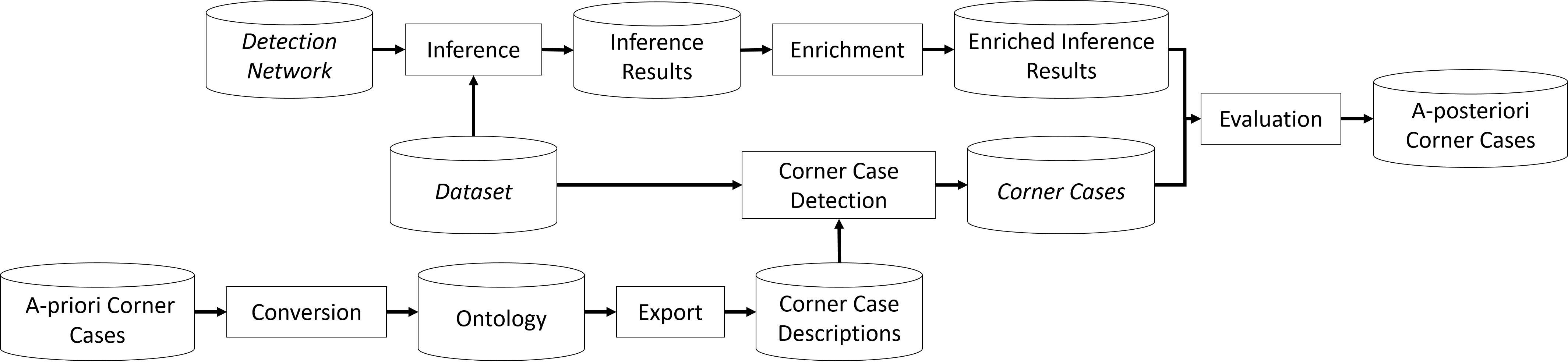}
           \captionof{figure}{The semi-automated corner case detection pipeline.}
           \label{fig:pipeline}
        \end{center}
    }]
}

\maketitle
\thispagestyle{empty}
\pagestyle{empty}

\begin{abstract}
In order to deploy automated vehicles to the public, it has to be proven that the vehicle can safely and robustly handle traffic in many different scenarios.
One important component of automated vehicles is the perception system that captures and processes the environment around the vehicle. 
Perception systems require large datasets for training their deep neural network.
Knowing which parts of the data in these datasets describe a corner case is an advantage during training or testing of the network. 
These corner cases describe situations that are rare and potentially challenging for the network.
We propose a pipeline that converts collective expert knowledge descriptions into the extended KI Absicherung ontology.
The ontology is used to describe scenes and scenarios that can be mapped to perception datasets.
The corner cases can then be extracted from the datasets. 
In addition, the pipeline enables the evaluation of the detection networks against the extracted corner cases to measure their performance.
\end{abstract}

\section{Introduction}
Automated vehicles and advanced driver assistance systems (ADAS) have evolved greatly in recent years. 
However, in order to get them on the road, it is of immense importance that the system is safe and robust. 
The perception system plays an important role here, as it is the foundation for further decisions of the overall system. 
Among other things, large amounts of data are needed for the perception system to work well enough. 
The selection of data is a challenging task, as it is still not understood what data and how much is needed to make a perception system fully functional.
But, in order for the system to be able to recognize everything in all contingencies, it is necessary to identify not only everyday situations, but also special cases that occur only rarely. 
Only if a perception system can also recognize and assign these situations correctly, it can be used reliably.

Identifying the right data for development and testing of a perception system is therefore an important task in automated applications.
Thus, a systematic way of finding this data must be established and existing datasets in the industry shall be reused in an efficient way. 
In the previous decades, perception data was collected in non consistent fashion regarding labeling and meta data. 
This makes the structured analysis of corner cases difficult and results in general deep neural network (DNN) metrics without understanding the data that was used for training and testing. 
This paper shall provide an overview of how corner cases can be described a-priori based on the latest research and existing datasets reused to analyze the DNN metrics a-posteriori. 

For this, the paper is structured as follows. 
In the second section, an overview of existing corner case definitions, methods to identify and classify corner cases and ontologies is given. 
The next section presents our approach on how to get from a-priori corner cases to a-posteriori corner cases of a dataset. 
We present the exemplary execution of the pipeline with an existing dataset in the fourth chapter and discuss advantages and disadvantages of the method.

\section{Related Work}
\label{sec:related_work}
Corner cases are of high interest during development, training and testing of DNNs. 
However, there is no common definition of them in the literature. 
Nevertheless, most of the various definitions share similar properties that characterize corner cases. 
Thus, first of all, it is important that these cases are rare. 
In \cite{KIDT_CCDescription} corner cases are described as data that occur infrequently, \cite{KIDT_CCCategorization_Camera} refers to situations that are unknown and have not been witnessed during training. 
\cite{KIDT_CCCategorization_CamLidRad} speaks of new cases or cases that deviate from the "normal".
A second important property of corner cases is that they have an impact.
They are described as critical \cite{KIDT_CCDescription}, unexpected \cite{KIDT_CCCategorization_Camera} or possibly dangerous situations \cite{KIDT_CCCategorization_CamLidRad} and as relevant for driving behavior \cite{KIDT_CCCategorization_CamLidRad}.
\cite{bolte2019_CCDef} summarizes these properties for automated driving corner cases in the following definition: "A corner case is given if there is a non-predictable relevant object/class in relevant location".
The reason why there is no common definition for corner case might be that what is considered as corner case depends on the application. 
In \cite{cc_trajectory}, the focus is on trajectory corner cases.
Here, the behavior of the different agents and their interaction is important to determine an anomalous case.
\cite{KIDT_CCDescription}, \cite{KIDT_CCCategorization_Camera} and \cite{KIDT_CCCategorization_CamLidRad} consider detection corner cases where challenges in the pure perception are relevant. \\ 

 \cite{KIDT_classification} introduces a classification of corner cases into different levels. 
 In camera based perception systems, the \textit{Pixel Level} consists of all corner cases with errors in the perceived data. 
 Here, they distinguish between a \textit{Local} and a \textit{Global Outlier}.
 A local outlier describes cases where only one or a few pixels deviate from expectations whereas a global outlier affects many or even all pixels.
 Local outliers can be caused by dead pixels or dirt, for example, while global outliers can be due to overexposure.
 In the \textit{Domain Level} all cases are summarized, which are caused by changes in the environment.
 Therefore, all corner cases related to a domain shift, e.g. due to weather conditions, different locations, etc., belong to this level.
Objects not seen during training are collected in the \textit{Object Level}.
 This can be a costumed person, an uncommon animal like a bear, or less commonplace items like wheelchairs or rollators.
 The next defined level is the \textit{Scene Level}. 
 Here, the objects under consideration are well known, however their recognition can be difficult.
 Either because there are many of these objects (\textit{Collective}), e.g. in a traffic jam or a demonstration, or because they are in a new context (\textit{Contextual}), e.g. a tree lying on the street.
Last but not least, there is the \textit{Scenario Level}. 
In contrast to the previous levels, a sequence of images/situations is considered and not just a single one.
Three cases can be distinguished: \textit{Risky}, \textit{Novel} and \textit{Anomalous Scenarios}. 
Risky scenarios are those that are well known but remain dangerous, such as overtaking a cyclist.
The opposite of these are novel scenarios. 
They are not known, but also do not pose a major hazard, e.g. accessing the freeway.
The most dangerous cases are combined in the anomalous scenarios. 
These are those where the situation is unknown and increases the probability of a critical situation, e.g. a person suddenly crossing the road.
In the order described here, the levels are ranked according to their detection complexity (from low to high).

These levels (introduced in \cite{KIDT_classification}) are then taken from \cite{KIDT_CCCategorization_CamLidRad}, further extended and summarized into layers.
For the scenario levels, the \textit{Temporal Layer} is introduced.
In the \textit{Content Layer}, domain, object and scene levels are summarized.
For the lowest level,  the \textit{Sensor Layer} is introduced. 
It shall cover all the cases from pixel level (camera related corner cases) extended by all cases caused by different sensor sources like LiDAR and RADAR.
Instead of dividing sensor cases into local and global outliers like \cite{KIDT_classification} did, \cite{KIDT_CCCategorization_CamLidRad} distinguishes between \textit{Hardware Level} and \textit{Physical Level}. 
Previous cases from pixel level are remapped to the two new levels. 
On hardware level, all issues due to hardware are located, e.g. dead pixel for camera, low voltage for RADAR or a broken mirror for LiDAR.
Physical issues like black objects for LiDAR, overexposure for camera or interference for RADAR are categorized in the physical level.
Additionally, \cite{KIDT_CCCategorization_CamLidRad} introduces a layer for all cases occurring due to the DNN called the \textit{Method Layer}.
Here, cases due to the topology, design or deployment are located.

\begin{table}[t!]
\caption{Classifications of corner cases used in pipeline.}
\begin{center}
\begin{tabular}{cccc}
\hline
\multicolumn{4}{c}{\textbf{Sensor Layer}}\\ 
\hline
\multicolumn{2}{c}{Physical Level}  & 
\multicolumn{2}{|c}{Hardware Level} \\
\hline
\multicolumn{1}{c}{Global Outlier} & 
\multicolumn{1}{|c}{Local Outlier} & 
\multicolumn{1}{|c}{Global Outlier} & 
\multicolumn{1}{|c}{Local Outlier}\\
\hline
\\
\hline
\multicolumn{4}{c}{\textbf{Content Layer}} \\
\hline
\multicolumn{1}{c}{Domain Level} & 
\multicolumn{1}{|c}{Object Level} & 
\multicolumn{2}{|c}{Scene Level}  \\ 
\hline
\multicolumn{1}{c}{} & 
\multicolumn{1}{c}{}  & 
\multicolumn{1}{|c|}{Collective} & 
\multicolumn{1}{c}{Contextual} \\
\hline
\end{tabular}
\end{center}
\label{tab:cc_classification}
\end{table}

In addition to the layer and level classification, \cite{KIDT_CCCategorization_CamLidRad} distinguishes between single- and multi-source corner cases.
Single-source corner cases are all cases that occur in the detection of one source. For example, the overexposure is a problem for the detection with camera. 
Multi-source corner cases occur on a different level of fusion.
Here, the interesting cases are those where the sources do not agree and after fusion a faulty result is produced.\\

Besides the corner case definitions and their classification, it is important to have a structured way to describe them.
To this end, an ontology can be used.
In an ontology, knowledge can be arranged systematically, value ranges can be defined and causalities between information can be created. 
Thus, an ontology has the advantage that knowledge is described formally and consistently.
Furthermore, it is possible to create rules with which additional knowledge can be interfered.
In the KI Absicherung\footnote[3]{https://www.ki-absicherung-projekt.de/en/} (\mbox{KI-A}) project, an ontology was developed to allow descriptions of single images (scenes) in a machine readable way. 
They use their ontology for two purposes. 
First, for labeling training and test data and second, for generating synthetic data \cite{kiaontologie}. 
There are many classes captured in the ontology that describe different parts of a scene.
This ranges for example from road and weather conditions to object and pedestrian descriptions with their appearance in addition to the light source parameters.
This makes it possible to define a scene with a wide range of details.
However, the \mbox{KI-A} ontology only allows the description of single scenes (single points in time) and not of complete scenarios (time series).

\section{Methodology}

\subsection{Corner Case Definition}
\label{sec:cc_definition}
In the context of the pipeline, the focus is on the perception corner cases only. 
Since there is no common definition of corner cases as described in Section~\ref{sec:related_work}, a perception corner case is defined here as follows: 
A situation that may lead to erroneous detections and, as a consequence, an unexpected situation during driving.

The introduced pipeline uses a similar classification of corner cases into layers and levels as described in Section~\ref{sec:related_work}.
For the formal description of corner cases, an extension of the \mbox{KI-A} ontology~\cite{kiaontologie} is implemented.
This extension will be described in later steps.
As this ontology only supports the descriptions of single scenes/time points and not of complete scenarios/time lines, the temporal and method layers are not considered within this paper. 
The layers and levels used can be found in Table~\ref{tab:cc_classification}. 
Section~\ref{sec:related_work} describes two different types of levels for the sensor layer. 
In this paper, the information is combined to have a physical or hardware problem with either local or global effects. 
This is to allow the most accurate categorization of these corner cases.

In addition, corner cases are mapped to the sensor source that causes them. 
Since a corner case can be caused by several sensor sources, all combinations of RADAR, camera (video) and LiDAR are possible. 
In the following, each combination of the sensor sources is grouped under the term RaVioLi (RADAR, Video, LiDAR).
The final classification of corner cases is based on whether the issue occurs before (single source) or after (multi source) sensor fusion. 

\begin{figure}[t!]
\centerline{\includegraphics[width=\columnwidth]{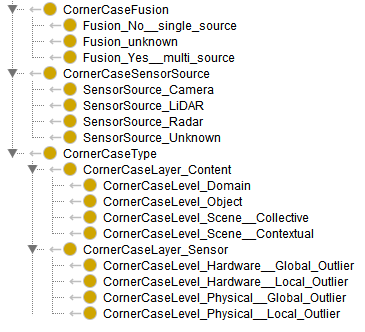}}
\caption{Classes added in ontology for corner case meta information.}
\label{fig:cc_ontology}
\end{figure}
\subsection{Corner Case Pipeline}
\label{sec:cc_pipeline}
In this section, the corner case detection and evaluation pipeline shown in Fig.~\ref{fig:pipeline} is described. 
The goal of this pipeline is to find and extract pre-defined corner cases from a given dataset, evaluate the performance of a DNN, and to extract specific corner cases to generate synthetic data. 
After manually collecting corner cases in an excel list (a-priori corner cases), they are described using the extended \mbox{KI-A} ontology~\cite{kiaontologie}. 
The ontology is used to automatically generate a generic list of potential corner cases with a set of metrics. 
In the context of the pipeline, metrics are sets of properties extracted from a scene description.
By mapping ontology classes to dataset specific properties, the extracted corner cases can be used to find real corner cases in the dataset.
Inference results by a DNN are then used to identify the a-posteriori corner cases.
The a-posteriori corner cases represent all situations in which objects were not detected.
In the following sections, each part of the pipeline is described in detail.

\subsubsection{A-priori Corner Case Collection} 
Many corner cases are well known to experts as they occur frequently or reach the limits of sensors. 
For use of those a-priori corner cases in the pipeline, the knowledge is maintained not using any specific corner case description language in an excel sheet. 
For each corner case, a description is added (e.g. overexposure) and causes that may lead to it (e.g. oncoming traffic at night or a low-standing sun). 
Furthermore, some meta information are tracked. 
First, there is the classification, i.e. in which layer and level the corner case is assigned to. 
It is possible to assign several layers and levels. 
For example, if rain is the cause of a corner case, it is a physical problem as the sensor itself has difficulties at handling the situation, nevertheless it is also a change in domain.
So both the physical level (sensor layer) and the domain level (content layer) can be assigned to the corner case.
As second meta information, the sensor source causing the corner case is added as RaVioLi (see Section~\ref{sec:cc_definition}). 
The last meta information is whether the corner case occurs before or after sensor fusion.
Exemplary rows of the Excel sheet can be found in Table~\ref{tab:cornercases}.\\

\begin{table*}[t!]
\caption{The implemented corner cases that are used for the experiments.}
\begin{center}
\begin{tabular}{m{0.01\textwidth} m{0.175\textwidth} m{0.285\textwidth}  m{0.05\textwidth} m{0.04\textwidth} m{0.07\textwidth}  m{0.175\textwidth}}
\cline{1-7} 
\textbf{No.} & \textbf{Description} & \textbf{Cause} & \textbf{RaVioLi} & \textbf{Source} & \textbf{Layer} & \textbf{Level} \\
\hline
1 & Camera overexposure & light of oncoming traffic at night & V & Single & Sensor & Physical - Global Outlier   \\
\hline
2 & High amount of vehicles & traffic jam/rush hour & R/V & Single & Content & Scene - Collective   \\
\hline
3 & Unusual persons & persons in wheelchairs & R/V/L & Single & Content & Object    \\
\hline
4 &Unusual objects & Traffic cones on street & R/V/L & Single & Content & Object  \\
\hline
& & & & & & Domain \\
\multicolumn{1}{l}{\multirow{-2}{*}{5}} & 
\multicolumn{1}{l}{\multirow{-2}{*}{Too many reflections}} & 
\multicolumn{1}{l}{\multirow{-2}{*}{rain}} & 
\multicolumn{1}{l}{\multirow{-2}{*}{L}} & 
\multicolumn{1}{l}{\multirow{-2}{*}{Single}} & 
\multicolumn{1}{l}{\multirow{-2}{*}{Content}} & Object   \\
\hline
& & & & & Sensor & Physical - Global Outlier \\
\multicolumn{1}{l}{\multirow{-2}{*}{6}} & 
\multicolumn{1}{l}{\multirow{-2}{*}{Attenuation back scattering}} & 
\multicolumn{1}{l}{\multirow{-2}{*}{rain}} & 
\multicolumn{1}{l}{\multirow{-2}{*}{R}} & 
\multicolumn{1}{l}{\multirow{-2}{*}{Single}} & Content & Domain   \\
\hline
7 & Multi path reflection & reflections on road & R & Single & Content & Object    \\
\hline
\end{tabular}
\label{tab:cornercases}
\end{center}
\end{table*}
\begin{figure}[t!]
\begin{subfigure}[b]{\columnwidth}
    \centerline{\includegraphics[width=\columnwidth]{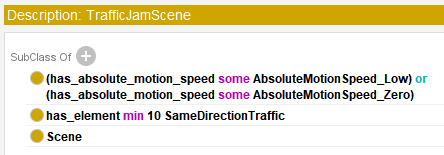}}
    \caption{Class for a scene with a traffic jam.}
\end{subfigure}
\begin{subfigure}[b]{\columnwidth}
    \centerline{\includegraphics[width=\columnwidth]{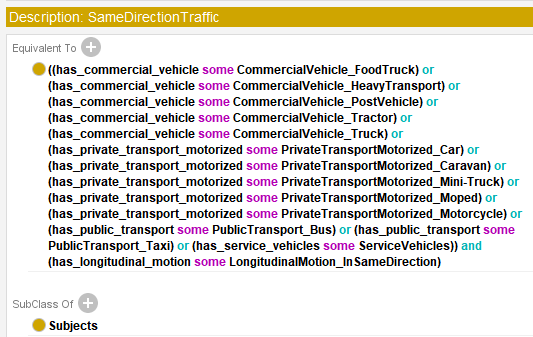}}
    \caption{Class for traffic driving in the same direction as the host.}
\end{subfigure}
\caption{Exemplary Description of Corner Case in Ontology.}
\label{fig:traffic_jam}
\end{figure}

Collecting corner cases in an excel sheet is an easy way to maintain the information in a human readable way. 
However, for adding more detailed information and to provide them in a well-structured way, it is not suitable. 
Therefore, the collected knowledge (in the latter called corner case meta information) is transferred into an ontology.
For this, the \mbox{KI-A} ontology is used.
But, to contain all the corner case meta information, this ontology needs to be extended and so a new sub-ontology is created. 
This sub-ontology consists of ontology classes for the different layers and levels, as well as classes for sensor source and fusion options.
The new added classes can be found in Fig.~\ref{fig:cc_ontology}.
Furthermore, a script adds for each row in excel sheet (corner case) and for each cause described for this corner case, a class in the ontology and connects it to the respective classification classes.
Thus after the script is executed, there is a class for each cause of a corner case having all its meta information assigned.

Additionally to the meta information, it is now possible to add concrete descriptions of the single corner cases in the ontology.
This is done by connecting a scene to a corner case description class. 
All information added to this scene should be necessary to receive the respective corner case. 
All other information not mentioned in the scene shall be arbitrary assignable without loosing the corner case property.
The description of scenes needs to be done manually. 
An exemplary description can be found in Fig.~\ref{fig:traffic_jam}. 
A traffic jam is described by at least ten vehicles driving in the same direction as the host vehicle, which has a low speed.

\subsubsection{Corner Case Extraction}
\label{sec:cc_extraction}
In a next phase of the pipeline, the corner cases described in the ontology shall be found in a dataset.
Therefore, the described scenes in the ontology are converted into metrics for the search in a dataset.
For this to work, the user needs to provide a mapping between the classes and properties used in the ontology and the information existing in the dataset.
Any information in dataset can be used for this: Labels, descriptions, CAN, map or any other meta information.
In case the dataset provides less information than specified in the ontology, several ontology classes or properties can be mapped to the same information in dataset. 
For example, if the ontology distinguishes between 'Food Truck', 'Heavy Transport', and 'Post Vehicle' but the dataset only provides a label for 'Truck', all three ontology classes can be mapped to the same truck class.
In the ontology, there are already value ranges set for some properties.
For instance, MotionSpeed\_Zero is a speed in range of $[0, 0.54]$ km/h.
Unless there are values provided in the excel sheet, the ones defined in ontology are used for the metrics.

Having this mapping from the user, a script can parse the ontology and convert each scene description for a corner case into a metric. 
All meta information described in the a-priori excel sheet are also extracted from ontology.
The extracted information - meta and metric information - are saved in a json file.

With this json file, a script browses the dataset and finds all data that matches the described metrics. 
This script depends on the used dataset and needs to be adapted accordingly if the dataset is changed. 
Therefore, the results may look different as dataset specific identifiers are used to describe what data matches the metrics. 
However, for all datasets, the data found by the script are the annotations that are predicted to be challenging for DNNs. 
Thus, these are the concrete a-priori corner cases extracted from a dataset derived from predefined expert knowledge.

\subsubsection{Dataset Evaluation on Neutral Network}
\label{sec:dataset_evalutation}
The comparison between annotations from the found corner cases in the dataset and the predicted objects requires a DNN.
The results of this comparison helps to identify gaps in the training data and stress test the network as the detected corner cases should only contain challenging scenes. 
Necessary steps for the dataset evaluation are on the one hand the inference of the dataset with the DNN and on the other hand the enrichment of those. 

The first step is to get the inference results for the selected subset of the dataset (validation or test).
It is necessary that there is a way to uniquely identify each object and its bounding box to ensure traceability between ground truth data and inference results.
Each predicated bounding box must be able to be associated with a bounding box in the ground truth data and its annotation information.
In addition, it must be ensured that each bounding box can also be assigned to the associated frame and to each scene. 

The next step is to enrich the inference results with all available information from the dataset (all labeled annotation data, meta data, scene and sample data, additionally generated data).
Additionally, every annotation is categorized into three different groups: false positive (FP), true positive (TP) and false negative (FN). 
True negatives (TN) are implicit in these numbers. 
Therefore, a matching algorithm is needed.
For simplicity, a simple euclidean distance matching algorithm (in 2D) with the Hungarian algorithm~\cite{Kuhn1955Hungarian} is performed between the ground truth data and the detected objects. 
After matching, every detected object is classified as one of the three categories (FP, TP, FN) and a corresponding flag is set in the enriched data. 
The enriched data is used to gain a better understanding of the object or to generate synthetic data at a later time. 

Once the inference results are enriched, the comparison between the inference and the detected corner cases can be performed.

\subsubsection{Evaluation of Corner Cases}
The last part of the pipeline combines the previous results to receive the a-posteriori corner cases.
We classify an a-priori corner case as an a-posteriori corner case if it is not detected by the DNN (being a FN).
This provides information about which a-priori corner cases are actually corner cases when considering a particular DNN.
The a-posteriori corner cases are related to the used DNN, since the results will change if a different network is used.

The results are used to generate statistics about the relation between a-priori and a-posteriori.
These statistics can provide insight about the performance of the network on each corner case.
As a result, it is possible to identify where the DNN is not yet performing well. 
However, the results can also be clustered into the corner case layers or levels and thus the network can be explicitly evaluated for specific corner case layers. 
For instance, if the performance on the physical layer is not sufficient, one can take a closer look at the corner cases of this layer.

\begin{figure}[t!]
\centerline{\includegraphics[width=\columnwidth]{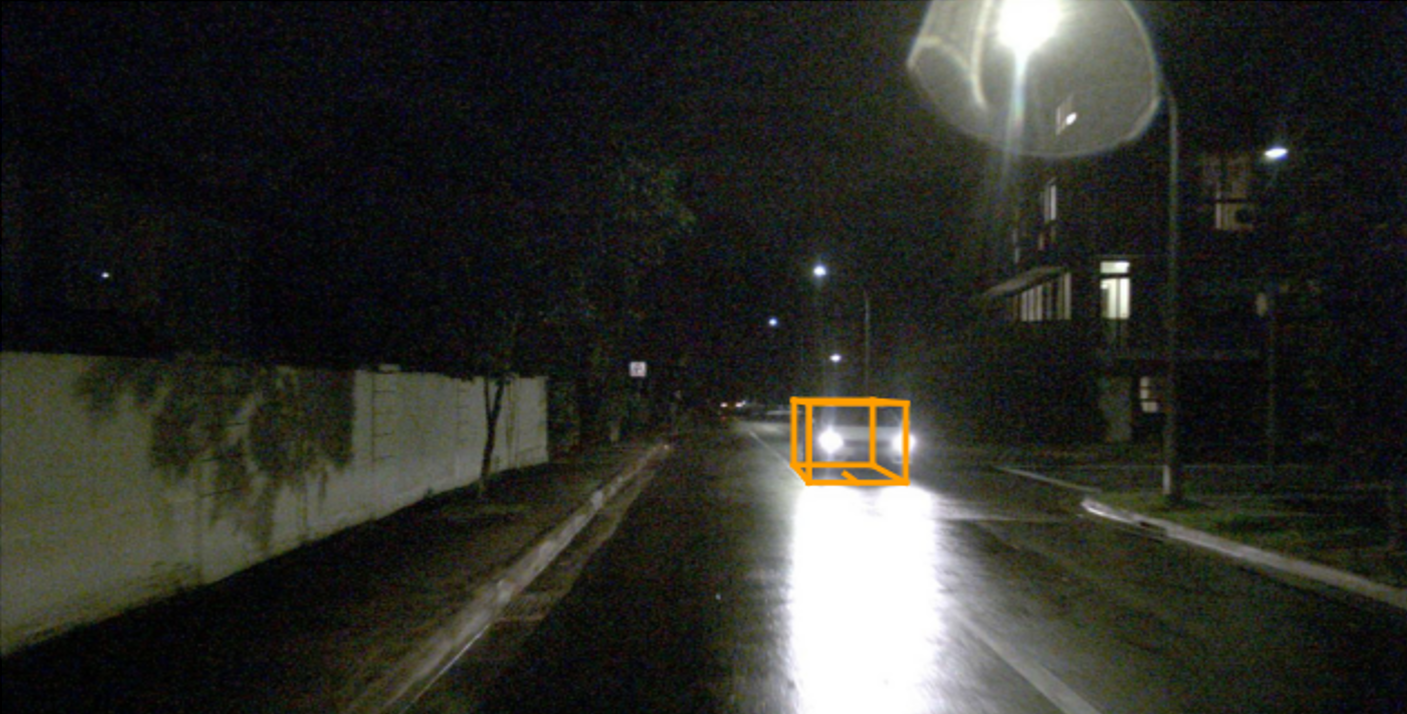}}
\caption{This image shows an oncoming car at night. This scene represents the overexposure corner case from Table~\ref{tab:cornercases}.}
\label{fig:cc_night_oncoming}
\end{figure}
\section{Experiments}
\label{sec:experiments}

\subsection{Dataset and Neural Network}
We used the nuScenes dataset~\cite{nuscenes2019} to perform the experiments.
There are 1000 scenes of 20 seconds length from Boston and Singapore containing 1.4 million camera images, 390~000 LiDAR sweeps, 1.4 million RADAR sweeps and 1.4 million object bounding boxes in 40~000 key frames.
In nuScenes, scenes are 20-second segments, samples are individual frames, and annotations are individual objects. 
The scenes are annotated at two Hz with 23 object classes and 3D bounding boxes. 
The dataset is split into two subsets: train with 850 scenes and validation with 150 scenes.
Also, nuScenes has an internal test set that is used to evaluate models for detection or tracking challenges.
However, this test set is not available for the public.

We use CenterFusion~\cite{nabati2020centerfusion}, which is a 3D DNN that utilize the RADAR and camera information to predict 3D bounding boxes.
It is one of the few networks that solely uses camera and RADAR data and has a working open source implementation. 
CenterFusion achieves a 0.449 nuScenes detection score (NDS) on the test set, which is a weighted sum of mean average precision (mAP), mean average translation error (mATE), mean average scale error (mASE), mean average orientation error (mAOE), mean average velocity error (mAVE) and mean average attribute error (mAAE).
In the scope of this paper, the network is used as a black box to generate the inference results and is not modified.

\subsection{Extracted Corner Cases}
\label{sec:exp-extracted_corner_cases}
As mentioned in Section~\ref{sec:cc_pipeline}, the corner case creation process of bringing the corner case into the ontology contains some manual work as the scene needs to be composed together by using available ontology components.
Due to complexity, seven selected corner cases are exemplary created and processed through the pipeline. 

\begin{figure}[t!]
\centerline{\includegraphics[width=\columnwidth]{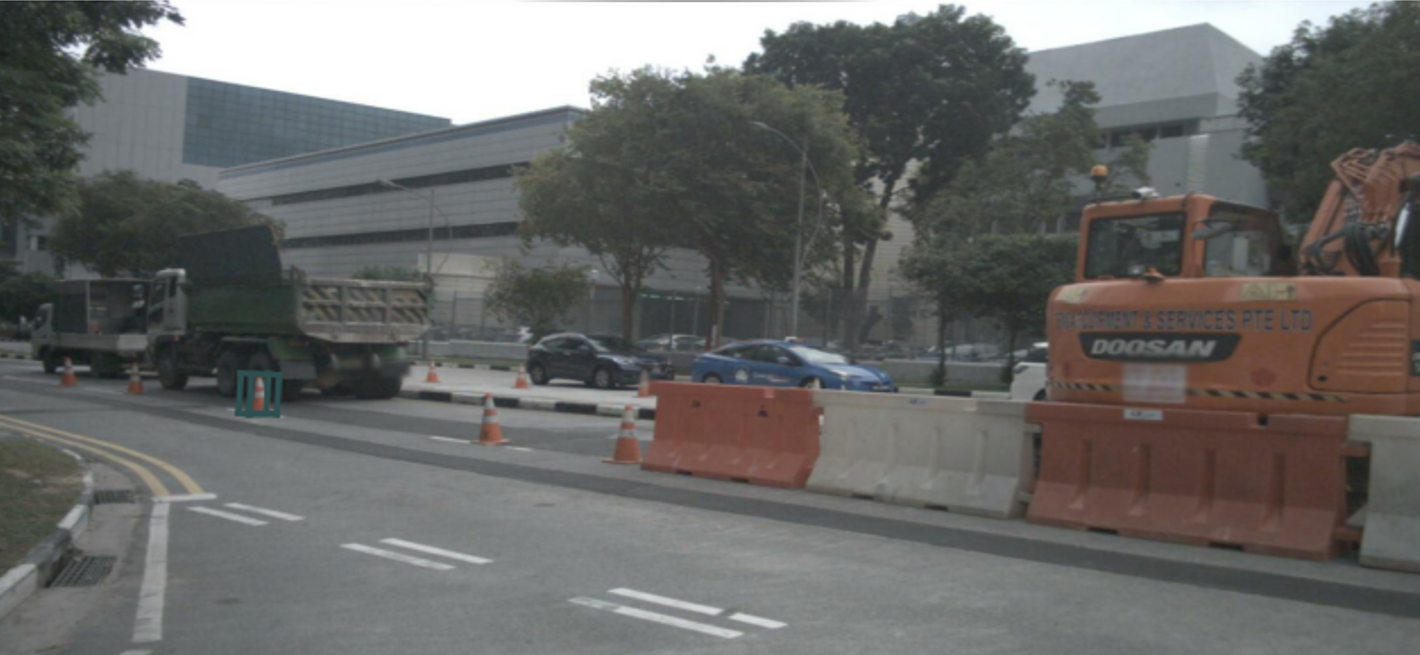}}
\caption{This image shows a construction area with traffic cones. This scene represents the "unusual objects" corner case from Table~\ref{tab:cornercases}.}
\label{fig:cc_traffic_cones}
\end{figure}
The corner cases that are used for these experiments are described in Table~\ref{tab:cornercases}.
They are selected to represent several modalities and scenes. 
The corner cases can be detected by their defined metrics.
However, as described in Section~\ref{sec:cc_extraction}, the metrics and values of the ontology need to be mapped to the nuScenes labels. 
Each value from the ontology used in the corner case requires a corresponding value in the nuScenes dataset. 
For instance, the object classes in the ontology are much more detailed than in nuScenes.
So, the detailed classes like 'Tractor', 'Food Truck', 'Heavy Transport', and 'Post Vehicle' need to be mapped on the simpler nuScenes class 'Truck'. 
In general, the detailed annotation process of nuScenes made mapping straightforward in most cases. 
However, nuScenes misses some basic parameters like weather or scene parameters. 
Domain values like weather conditions (e.g. rain, sun elevation) or daytime (e.g. night or day) are difficult to map accurately as the information are only contained in the scene descriptions.
Hence, we need to use these descriptions which contain a short informal text of what happens in the scene. 
There is no specific structure or constraint on how the description looks like, which makes it difficult to find any specific information. 
Filtering by particular values is not enough, as there are many misspellings or shorthanded descriptions.
When filtering for rain, the result can lead to a scene description that contains either heavy rain or no rain. 
By using natural language processing tools such as tokenization~\cite{tokenization} and named entity recognition~\cite{ner}, we can achieve better accuracy than by searching for single words.

The extracted corner cases contain around 32~000 out of a total of 122~778 annotations, 2025 out of a total of 6019 samples and 83 out of a total of 150 scenes of the validation set.
So, about 27\% of all annotations are already covered by seven corner cases.
This is due to the fact that two corner cases search for rain, which can only be found in the informal scene descriptions. 
As a result, all frames of a scene with rain are classified as corner cases. 
Furthermore, false positives can occur if the parameter range is not precise enough.

\begin{figure}[t!]
\centerline{\includegraphics[width=\columnwidth]{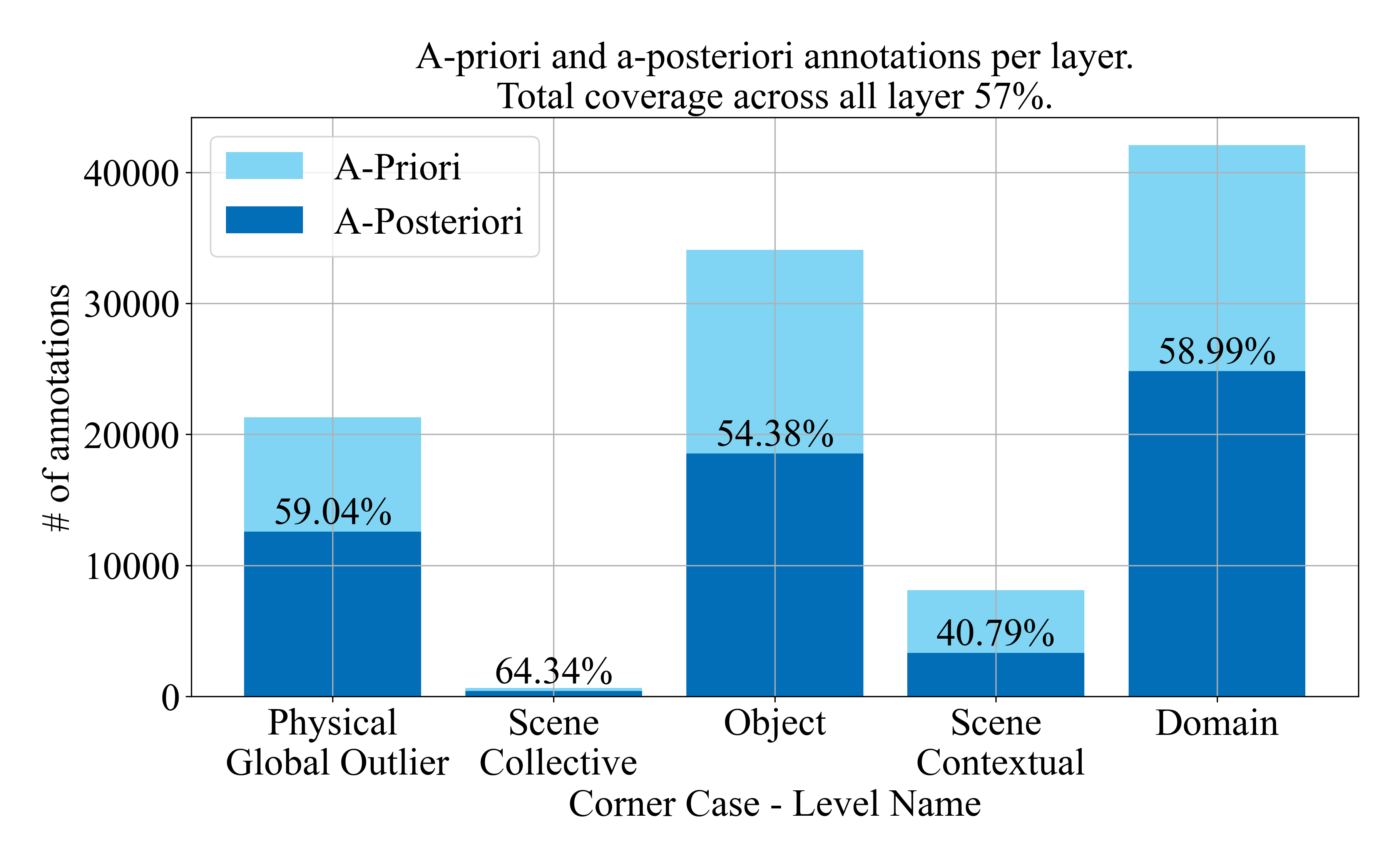}}
\caption{Number of annotations per layer.}
\label{fig:num_ann_layer}
\end{figure} 

In Fig.~\ref{fig:cc_night_oncoming} and Fig.~\ref{fig:cc_traffic_cones}, two extracted corner cases from the nuScenes dataset are shown. 
Fig.~\ref{fig:cc_night_oncoming} shows oncoming traffic at night producing overexposure for the camera.
The orange box is the ground truth 3D bounding box for car.
This can be a challenging scene for a camera because the dimensions of the car are barely visible.  
The image in Fig.~\ref{fig:cc_traffic_cones} contains traffic cones, which is also a challenging scene. 
For the camera, the pattern of red and white stripes is unusual and could therefore result in wrong detections or classifications. 
For radars, the  shape of the pylon is difficult as it leads to uncommon reflections.
Thus for the perception system it is challenging to determine the drivable area which is temporarily restricted by the traffic cones.

This last step summarized the first part of the pipeline by proceeding from collecting a-priori corner cases, importing the corner cases to an ontology to exporting and finding the corner cases in real world datasets. 
Additionally, we extended the pipeline even further to evaluate these corner cases with a DNN.
The framework is then able to determine whether an a-priori corner case is also an a-posteriori corner case for this explicit DNN.
The details of the last part of the pipeline is discussed in the next sections.

\subsection{Inference and Enrichment}
The next step is to receive the inference results of the DNN. 
Therefore, we use CenterFusion to predict 3D bounding boxes for the validation dataset of nuScenes. 
CenterFusion already converts its results to the nuScenes format so that they can be processed directly by the nuScenes developer kit\footnote{https://github.com/nutonomy/nuscenes-devkit}.
The standard evaluation of nuScenes contains a class-wise center-based matching on the xy plane of the ground truth and the predictions. 
A match is achieved when the distance of two center points is smaller than 50 centimeters.  
However, no information about false positives, false negatives and true positives is saved during this process.
We add the assignment of these categories to the algorithm as described in Section~\ref{sec:dataset_evalutation}.

\begin{figure}[t!]
\centerline{\includegraphics[width=\columnwidth]{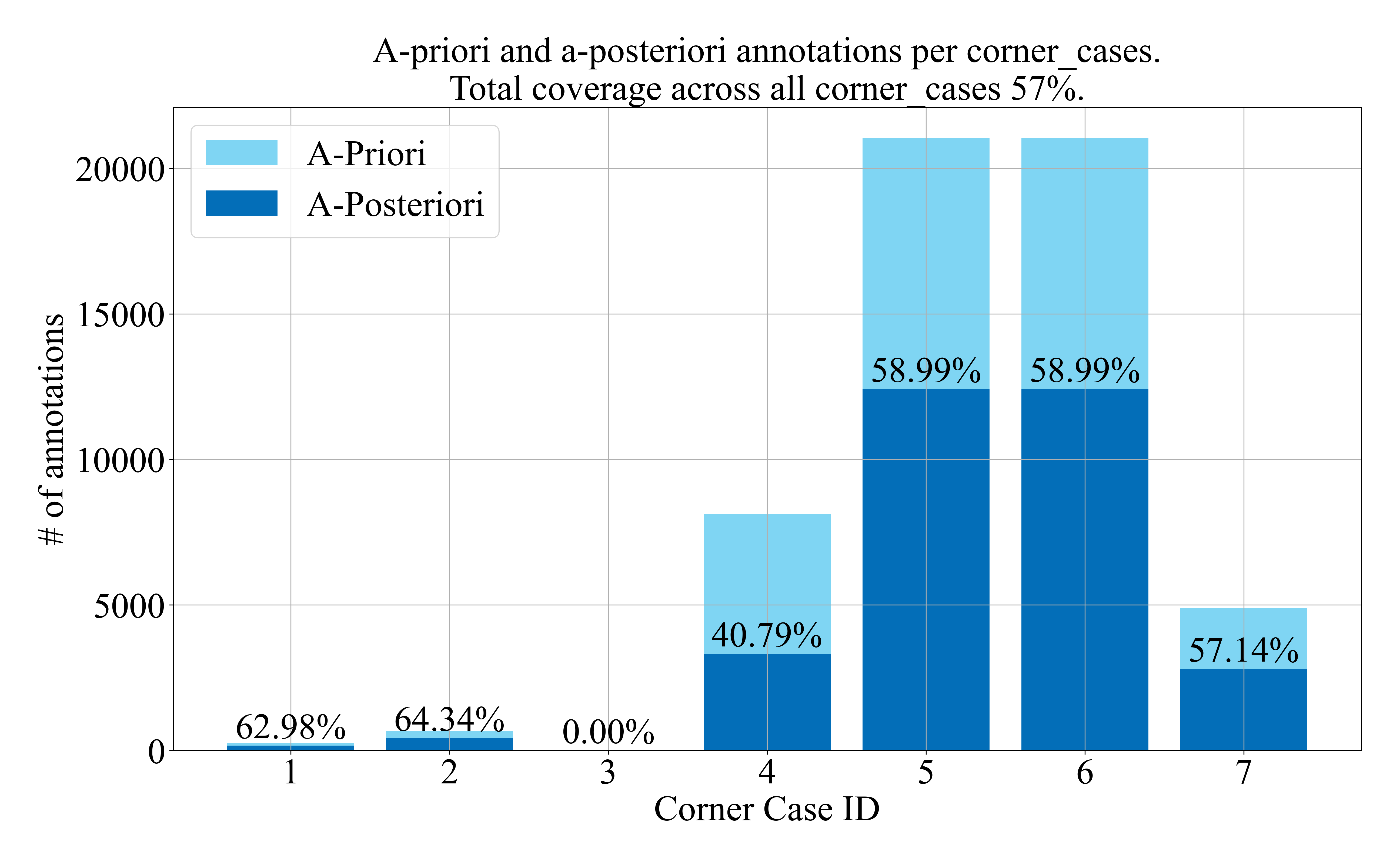}}
\caption{Number of annotations per corner case.}
\label{fig:num_ann_cc}
\end{figure}
\subsection{A-Priori vs. A-Posteriori}
At this point, the extracted corner cases and the inference results from nuScenes are available. 
We evaluate the a-priori corner cases in terms of whether they are false negatives or not (e.g. being an a-posteriori corner case or not). 
If a scene contains false negatives, it is considered an a-posteriori corner case. 
After deriving the a-posteriori corner cases from the a-priori corner cases, statistics of the relation between both can be created. 
These statistics are shown in Fig.~\ref{fig:num_ann_layer} and Fig.~\ref{fig:num_ann_cc}.   
Following questions can be answered when analyzing these statistics: 
\begin{itemize}
    \item Are there corner case layer specific issues causing the DNN to not work properly?
    \item Are there specific corner cases that the network is performing not well on? 
    \item Are the parameters for the a-priori corner case detection not set well because we have a high coverage?
    \item Are corner cases completely missing in the data? 
\end{itemize}
The numerical basis for both diagrams is the same, but the data is summarized differently for each diagram. 
In Fig.~\ref{fig:num_ann_layer}, the light blue bar displays the a-priori corner cases and the dark blue bar displays the a-posteriori corner cases. 
Evaluating the corner cases on layer level can give insights on how well a network works on a specific layer. 
For instance, if there are many a-posteriori corner cases in the physical layer, there could be a problem in the sensor itself or limitations at the physical level of the sensor.
The results in Fig.~\ref{fig:num_ann_layer} show that on average 57\% of corner cases are real corner cases for the DNN. 
As CenterFusion is not performing well on the nuScenes dataset with a 0.449 NDS on the test set, an average of 50\% coverage is expected. 
The amount of a-posteriori corner cases is closely related to the performance of the used DNN. 
Using a better network would result in less a-posteriori corner cases.
The different sensor modalities used influence the results as well. 
CenterFusion uses camera and RADAR, but no LiDAR data. 
When comparing the scores for the nuScenes challenge, the best scores, ranging from 0.73-0.77 NDS, were obtained from networks using LiDAR data such as in~\cite{liu2022bevfusion, mgtanet, kim20223ddf, wang2023dsvt}.

In Fig.~\ref{fig:num_ann_cc}, the relations of the a-priori and a-posteriori corner cases are shown for each corner case.
It is noticeable that no data was found for the wheelchair corner case (corner case number three). 
This shows that the pipeline can also determine whether certain corner cases are available in the data.
The information about missing corner cases can be used to optimize the collection of data. 
Also, as already described in Section~\ref{sec:exp-extracted_corner_cases}, the corner cases five and six have the same amount of corner cases because the only value that could be used to find these corner cases was the keyword rain (see Table~\ref{tab:cornercases}).
This also emphasizes that several corner cases, despite their different descriptions, can have the same cause.

Increasing the amount of defined corner cases and using a better DNN would result in a more meaningful statistic representing the performance of the network on the specific dataset.

\section{Conclusion}
We presented a novel pipeline to extract corner cases described in an excel list almost automatically from a real world dataset. 
The corner cases in the excel list are described using an ontology. 
The ontology is then used to export the corner cases to any dataset by mapping ontology values onto dataset specific values (map class labels of ontology onto nuScenes class labels). 
We can then extract all the defined corner cases from the dataset and evaluate them with a DNN.
The results give us insights of whether a corner case is considered a corner case and how well a network performs on those corner cases.

Our proposed method can be used to collect and define a set of corner cases, import them into the ontology, export and map the corner cases from the ontology onto a dataset and then extract the exported corner cases. 
The corner cases in the ontology can be used for any dataset if there is a mapping between ontology values and dataset values.
This shows that it would be a major improvement if different datasets would share the same labels e.g. based on the classes defined in the KI-A ontology.
In addition, DNNs can be evaluated against the corner cases to check performance in challenging situations. 
Also, the extracted corner cases from the datasets can be used to generate synthetic data with variation, resulting in higher variance for the DNN to improve network performance.

The \textit{Method} and the \textit{Temporal layer} are currently not considered.
A next step would be to integrate the two layers into the ontology and pipeline.
Especially the temporal layer would add an important set of corner cases.
Moreover, another step is to add a branch to the pipeline that is able to transfer these corner cases directly from the ontology to a simulation in a semi-automated way to generate synthetic data.
Generating synthetic data is a great way to extend the testing parameter range and to get data that would not get recorded in real world data collections.

\section*{ACKNOWLEDGMENT}
The research leading to these results is funded by the German Federal Ministry for Economic Affairs and Climate Action within the project “KI Data Tooling – Methoden und Werkzeuge für das Generieren und Veredeln von Trainings-, Validierungs- und Absicherungsdaten für KI-Funktionen autonomer Fahrzeuge". The authors would like to thank the consortium for the successful cooperation.



\bibliographystyle{IEEEtran}
\bibliography{main} 

\end{document}